\documentclass[letterpaper, 10 pt, conference]{ieeeconf}  

\IEEEoverridecommandlockouts                              

\usepackage{amsmath, amssymb}
\usepackage{graphicx}
\usepackage{cite}
\usepackage{xcolor}

\title{\LARGE \bf 

An analysis of sensor selection for fruit picking with suction-based grippers 

\author{Eva Krueger$^{1}$, Marcus Rosette$^{1}$, and Joseph R. Davidson$^{1}$}
\thanks{This research was supported in part by the Washington Tree Fruit Research Commission.}
\thanks{$^1$Collaborative Robotics and Intelligent Systems (CoRIS) Institute, Oregon State University, Corvallis OR 97331, USA {\tt\footnotesize \{kruegeev,rosettem,joseph.davidson\}@oregonstate.edu}}
}

\begin{document}

\maketitle

\begin{abstract}
Robotic fruit harvesting often fails to reliably detect whether a fruit has been successfully picked, limiting efficiency and increasing crop damage. This problem is difficult due to compliant fruit and grippers, variable stem attachment, and occlusions in orchard environments. Prior work has explored vision-based perception and multi-sensor learning approaches for pick state estimation. However, minimal sensor sets and phase-dependent sensing strategies for accurate pick and slip detection remain largely unexplored. In this work, we design and evaluate a multimodal sensing suite integrated into a compliant suction-based apple gripper. Our approach is unique because it identifies which sensors are most informative at different phases of the pick, enabling predictive detection of failures before they occur. The contributions of this paper are a phase-dependent evaluation of multimodal sensors and the identification of minimal sensor sets for reliable pick state classification. Experiments in a real apple orchard show that Random Forest and Multilayer Perceptron classifiers detect successful picks and impending failures with over 90\% accuracy, and Random Forest predicts pick/slip events within 0.09 s of human-annotated ground truth.

\end{abstract}

\section{Introduction}
\label{sec:intro}

Robotic harvesting of high-value tree fruit, such as fresh market apples and pears, is increasingly important due to labor shortages and rising production costs~\cite{davidson_2020, zhao_2011, senden_2022}. Despite advances in fruit detection, localization, and motion planning, reliably determining whether a fruit has been successfully detached remains a critical challenge \cite{velasquez_2022}. Robots often lack feedback during the final picking stages, leading to failed picks, wasted motion, or damage to fruit and trees.

Accurate real-time pick state classification -- detecting whether a fruit is securely held, slipping from the grasp, or has been successfully detached -- is essential to address these issues. This classification is particularly difficult because fruit are physically attached via stems, both fruit and grippers are compliant, and orchard environments introduce variability in fruit size, orientation, and occlusion from leaves or branches \cite{senden_2022, xiong_2020, sytsma_2025}.

Prior work has explored vision-based perception, local sensing, and learning-based pick state estimation \cite{senden_2022, velasquez_2022, walt_2025}. Vision-only approaches struggle during final grasp phases due to occlusion and short, noisy sensing (i.e. depth), while existing multi-sensor and learning-based systems rarely evaluate which sensors are truly necessary, or their performance under realistic orchard conditions. As a result, minimal sensor sets for reliable grasp detection and phase-dependent sensing strategies remain largely unexplored.

\begin{figure}
    \centering
    \includegraphics[width=0.9\linewidth]{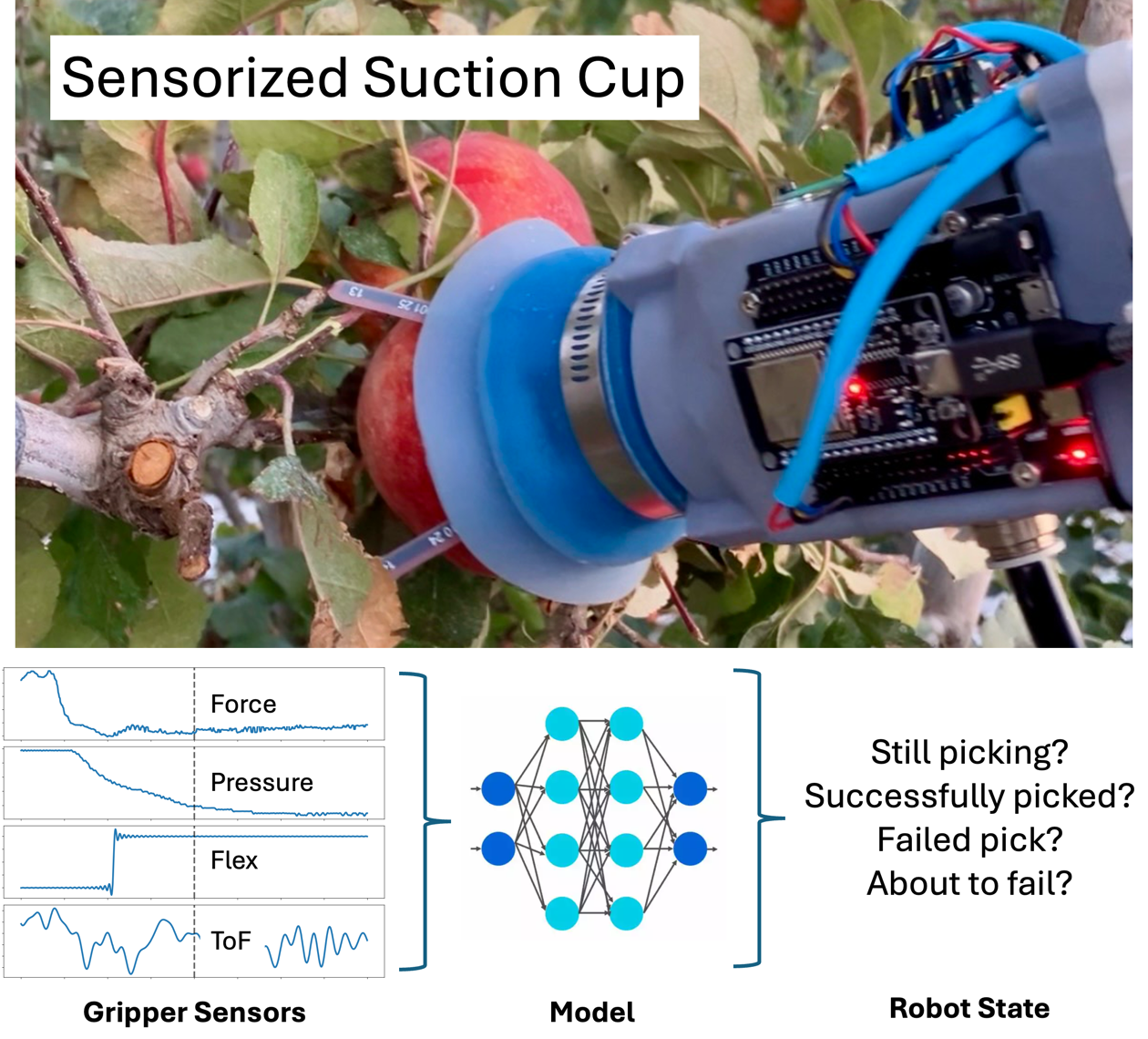}
    \caption{Pick state classification workflow. A compliant suction-based gripper interacts with an apple while embedded sensors (force, flex, pressure, and ToF) capture the picking dynamics. Sensor signals are fed into a machine learning model (Random Forest or neural network), which predicts the pick state in real time as \emph{picking}, \emph{pre-failure}, \emph{picked}, or \emph{failed pick}.}

    \label{fig:graphical_abstract}
\end{figure}

In this work, we design and evaluate a multimodal sensing suite for a compliant suction-based apple harvesting gripper. We analyze the contributions of force, strain/flex, vacuum pressure, and time-of-flight sensors to identify different grasp phases (Fig.~\ref{fig:graphical_abstract}), compare both Random Forest and Multi-Layer Perceptron (MLP) classifiers, and identify minimal sensor combinations that achieve reliable detection of both impending failures and successful picks under real orchard conditions. \textit{Our primary contributions are i) phase-dependent evaluation of multimodal sensors for pick state classification; and ii) the identification of minimal sensor sets that maintain reliable detection under realistic conditions}. By addressing these open challenges, this work enables more efficient, reliable, and deployable robotic fruit harvesting systems.

\section{Related Work}
\label{sec:related}

Early work on grasp and slip detection in agricultural manipulation relied on threshold-based force heuristics and analytic models to infer grasp quality \cite{zhao_2011}. More recent approaches increasingly employ data-driven methods to detect slip, loss of contact, or pick failure during manipulation \cite{velasquez_2022, walt_2025}. Across this literature, a consistent finding is that accurate, real-time pick state estimation is essential for enabling corrective actions, reducing failed picks, and improving harvesting efficiency under field conditions.

\subsection{Limitations of Vision-Based Approaches}

Vision-based perception is central to fruit detection and localization, but camera-only approaches exhibit fundamental limitations during the final approach, contact, and detachment phases of harvesting. Occlusion from foliage and stems, near-field depth noise, lighting variability, and reflective fruit surfaces degrade visual reliability at grasp distance \cite{senden_2022, xiong_2020}. More importantly, vision alone cannot reliably detect micro-slip, compliant deformation, or the moment of stem abscission, leaving many systems unaware of whether a pick has succeeded or failed \cite{davidson_2020, zhao_2011}. These limitations motivate the integration of local sensing at the end-effector to provide direct feedback during the most failure-prone phases of manipulation.

\subsection{Local Sensing for Abscission and Slip Detection}

A wide range of local sensing modalities has been explored for detecting grasp stability and pick outcomes in agricultural robotics. Commonly used sensors include force/torque (F/T) sensors, tactile or strain sensors, vacuum pressure sensors, and proximity sensors such as infrared or time-of-flight (ToF). Each modality captures different physical aspects of the interaction between the gripper and fruit.

Force and torque sensing provides direct measurements of load transfer and has been used to infer slip and successful detachment events \cite{dischinger_2021}. Vacuum pressure sensors reflect seal integrity in suction-based grippers and are effective for detecting partial attachment, leakage, or loss of suction \cite{velasquez_2024}. Flex and strain sensors embedded in compliant structures capture deformation due to contact changes or incipient slip, while proximity sensors can detect relative motion of the fruit away from the gripper during failure events \cite{Hemming2025Multisensor, dischinger_2021, walt_2025}. Prior work consistently shows that no single sensing modality is sufficient to robustly detect all grasp phases and failure modes in isolation \cite{velasquez_2022, davidson_2020}. To address this limitation, many recent systems integrate multiple local sensors and employ sensor fusion to infer pick state.

\subsection{Learning-Based Pick State Estimation in Agriculture}

Machine learning methods are widely used to interpret noisy, heterogeneous sensor data in agricultural robotics. Learning-based approaches map multimodal sensor streams to discrete pick states or success probabilities, enabling predictive detection of failure and real-time verification of pick success \cite{Hemming2025Multisensor, walt_2025}. Across a variety of crops and gripper designs, multi-sensor fusion has been shown to improve grasp reliability compared to single-sensor or vision-only systems \cite{hua_2025, sytsma_2025}. However, much of the existing work emphasizes overall detection accuracy rather than analyzing the relative contribution of individual sensors or identifying minimal sensor configurations suitable for practical deployment.

Random Forest (RF) classifiers are frequently adopted due to their robustness, low computational cost, and interpretability, making them well suited for real-time deployment \cite{walt_2025, liu_2024}. Multilayer perceptrons (MLPs) and related neural architectures have also been applied to model nonlinear relationships across sensor modalities and time \cite{li_2024}. Similar learning-based sensor fusion approaches have been successfully applied to agricultural sensing problems beyond harvesting, including crop monitoring and phenotyping \cite{zheng_2021, waqas_2025}. Despite these advances, comparative analysis of model behavior and sensor relevance under realistic orchard conditions remains limited.

\subsection{Gaps and Open Problems}

Despite substantial progress, several important gaps remain. First, few studies systematically evaluate which sensing modalities are necessary or redundant, leaving open questions about minimal viable sensor sets that balance performance with cost, weight, and mechanical complexity. Second, explicit real-time detection of fruit abscission is rarely addressed, even though it is critical for downstream decision-making such as terminating pull motions or initiating fruit transfer to a storage container. Finally, many sensorized grippers and learning-based grasp classifiers are validated primarily in laboratory or semi-controlled environments, with limited evaluation under real, commercial orchard conditions.

Building on prior work in local sensing and learning-based grasp estimation, the present study focuses on sensor selection and pick state classification for a compliant, suction-based apple harvesting gripper. By analyzing phase-dependent sensor relevance, comparing Random Forest and MLP classifiers, and evaluating performance under real orchard conditions, this work addresses key open challenges in practical robotic fruit harvesting.

\begin{figure*}
    \centering
    \includegraphics[width=0.9\linewidth]{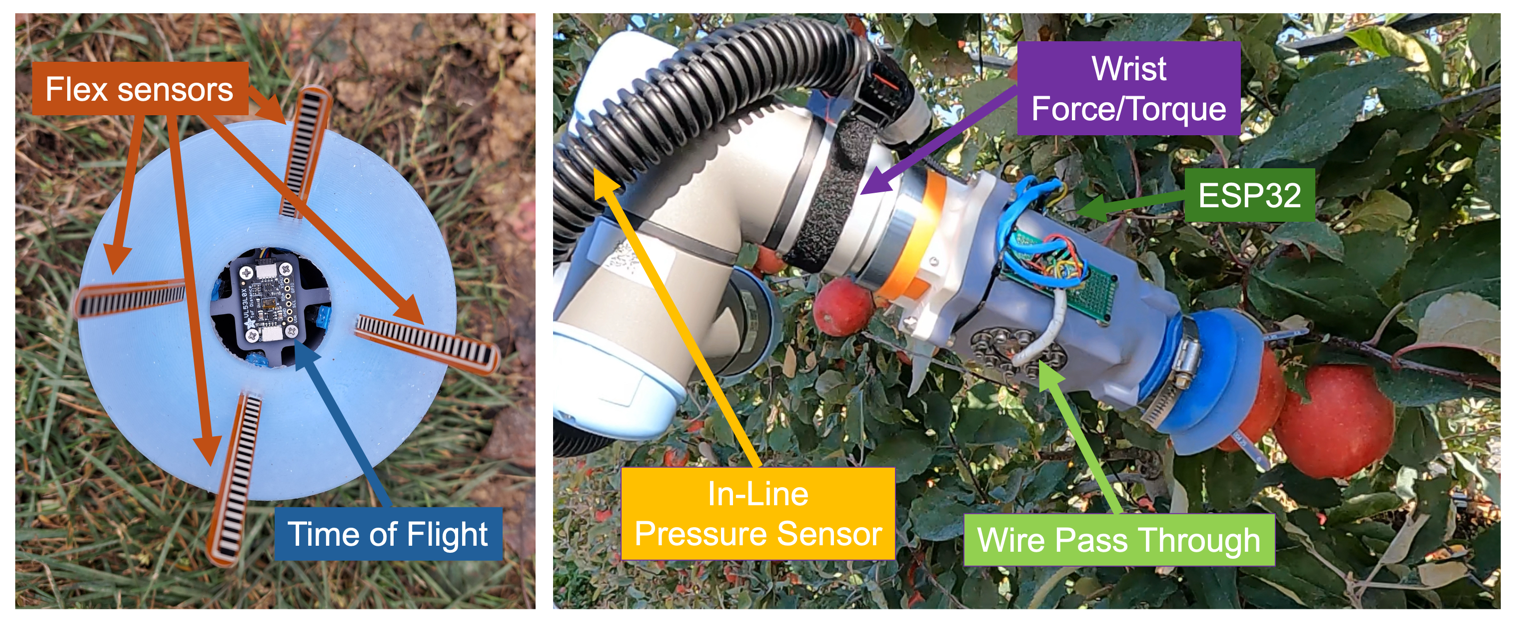}
    \caption{Compliant suction-based gripper with integrated sensing. \emph{Left:} Top-down view of the gripper showing the suction cup and embedded sensors, including vacuum pressure, flex, and time-of-flight (ToF) sensors. \emph{Right:} Gripper during an apple picking operation, illustrating compliant contact and suction engagement during stem abscission.}
    \label{fig:gripper_and_sensors}
\end{figure*}

\section{Methods}
\label{sec:experimentation}

This work investigates the use of local sensing in a compliant suction gripper for robotic apple harvesting. Experiments were conducted with a UR5e manipulator (Universal Robots, Odense, Denmark) performing repeated pick attempts in a commercial apple orchard in Prosser, Washington (U.S.A.), under natural conditions including variable lighting, occlusion, and fruit pose variability. Supervised learning models, including Random Forest and an MLP, were applied to classify grasp outcomes from the collected sensor data.

\subsection{Gripper Design}

The end-effector is a custom, compliant suction-based gripper, building on prior work in multi-sensor suction end-effectors \cite{Hemming2025Multisensor}. It employs a soft pneumatic suction cup molded using Dragonskin 10 (Smooth-On, Inc., Macungie, PA, USA) that accommodates variability in fruit size, geometry, and orientation while minimizing bruising. Passive compliance from the deformable cup enables stable engagement under minor misalignment due to interference from foliage or branches. A central vacuum channel generated by a piCOMPACT 23 SMART ejector (Piab, Danderyd, Sweden) provides attachment force, while the surrounding compliant structure maintains continuous contact during pulling and twisting motions required for stem abscission.

Multiple sensors are integrated at the gripper's contact interface. Four resistive flex sensors (Spectra Symbol FS-L-0055-253-ST) measure deformation from objects near the cup (such as apples, leaves, twigs, etc.). A time-of-flight sensor (STMicroelectronics VL53L0X) detects fruit distance from the suction cup. The manipulator's integrated 6-axis force/torque (F/T) sensor captures interaction forces. Sensor wires are routed through the vacuum enclosure using a custom cable pass-through. All gripper sensors are sampled by an ESP32 microcontroller and streamed to the robot control computer via ROS 2, where they are synchronized with robot pose and F/T measurements for analysis. An overview of the gripper design and sensor placement is shown in Fig.~\ref{fig:gripper_and_sensors}.

\begin{figure*}
    \centering
    \includegraphics[width=1.0\linewidth]{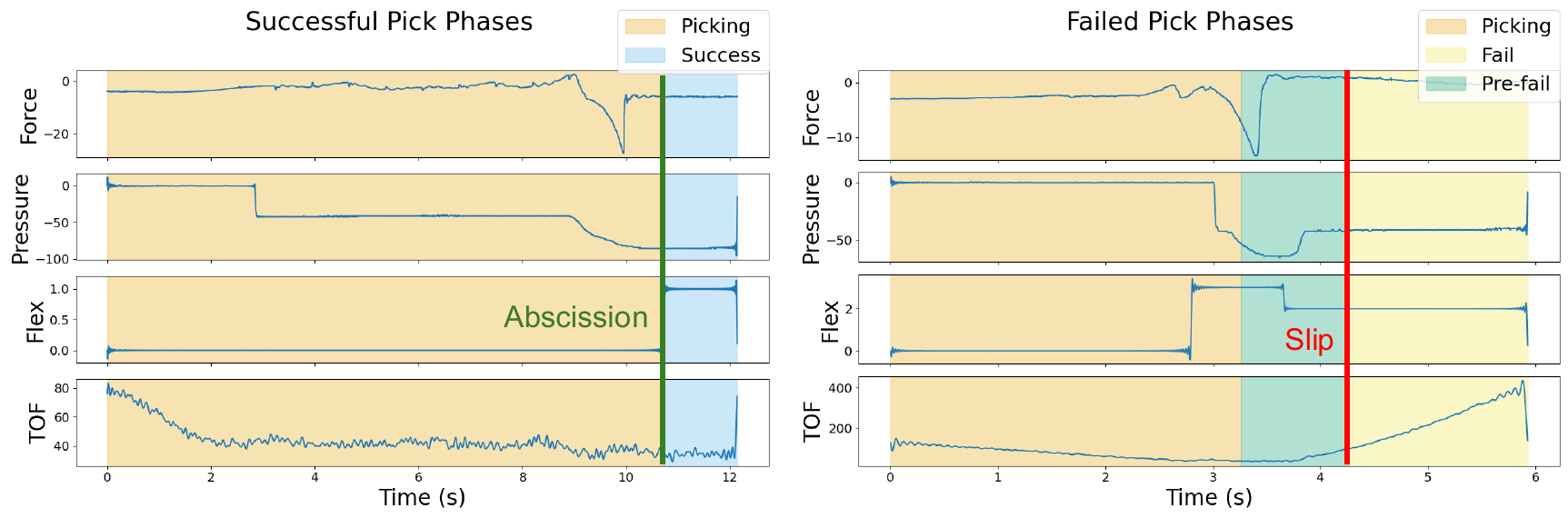}
    \caption{Definition of pick states used for time-series labeling. Sensor data are segmented into \emph{picking}, \emph{pre-failure}, \emph{picked}, and \emph{failed pick} states relative to grasp onset and abscission or slip events. If an apple is grasped, it is in the picking phase. If it starts to slip, it transitions to the pre-failure, then the failed pick phase. If it does not slip, it transitions eventually to the picked phase, after abscission.}
    \label{fig:grasp_states}
\end{figure*}

\subsection{Data Collection and Preprocessing}

As in previous work, the suction cup is initially aligned using closed-loop control based on local sensors~\cite{Hemming2025Multisensor, krueger_2025}. In this study, all trials begin after the suction has engaged the fruit, so the focus is on the pick dynamics after a successful grasp. Each trial starts with the fruit secured and the manipulator initiating the lifting or detachment motion, enabling consistent analysis of pick state evolution—including pre-failure and slip events. A total of 83 pick attempts were recorded (72 successful, 11 failed), with videos from multiple cameras observing the scene captured in ROS~2 bag files for annotation. Ground-truth times of picking/abscission or slip events were recorded independently by four human annotators observing the postprocessed videos. The average standard deviation between annotations for the 4 subjects was 0.16 seconds.

Sensor data were labeled into four classes: \emph{picking}, \emph{pre-failure} (1\,s before slip), \emph{picked}, and \emph{failed pick}. These labels (shown in Fig.~\ref{fig:grasp_states}) were chosen because they are informative for downstream manipulation decisions. Detecting whether a pick has completed successfully or failed (picked/failed pick) enables the system to decide whether to retry the action or mark the apple as harvested. Detecting an imminent failure (pre-failure) allows the system to intervene proactively, e.g. by retrying the pick, pausing the motion, or increasing suction. 

Data streams were cropped to start at grasp onset, detected via a vacuum pressure threshold. Force signals were denoised using median filtering; flex signals were normalized and smoothed; pressure and time-of-flight signals required minimal preprocessing. All streams were synchronized and resampled to a common length. The dataset was split into training (80\%), validation (10\%), and testing (10\%) sets at the trial level. To improve robustness and mitigate class imbalance, time-series data were augmented by adding per-channel Gaussian noise:
\begin{equation}
\sigma \sim \mathcal{U}\!\left(0, \alpha (\max(x)-\min(x))\right), \quad x_\text{aug} = x + \mathcal{N}(0, \sigma^2),
\end{equation}
where $\alpha$ defines the maximum augmentation percentage. Failed trials were augmented seven times, successful trials once, preserving temporal structure.

\subsection{Classification Methods}
\begin{figure*}
    \centering
    \includegraphics[width=1.0\linewidth]{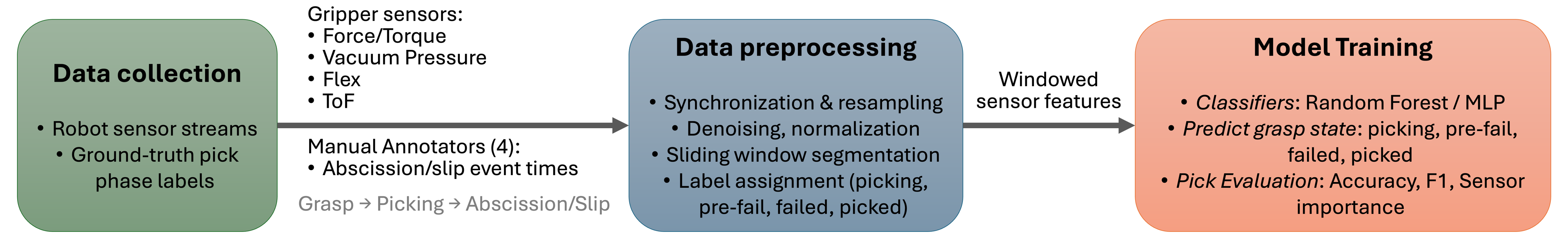}
    \caption{Pick state classification workflow. Raw sensor streams are collected from the gripper during picking. Ground truth labels are created from manual annotations (4 independent annotators). Data are preprocessed, synchronized, and segmented into sliding windows, then used to train Random Forest and MLP classifiers. Model outputs are evaluated against ground-truth annotations, and feature importance is computed to assess sensor contributions.}
    \label{fig:model_architecture}
\end{figure*}

Pick state classification was performed using fixed-length sliding windows of 5 time steps with a stride of 5, with each window labeled according to the pick state at its final time step. A Random Forest classifier with 100 decision trees was used as a baseline, enabling fusion of heterogeneous sensor inputs and computation of feature importance. The MLP was implemented using the scikit-learn \texttt{MLPClassifier}, and trained on standardized features, with scaling parameters fit on the training set and applied to validation and test data. The network consisted of two fully connected hidden layers (150 and 50 units) with ReLU activations. Training was performed for up to 200 epochs with mini-batch optimization and early stopping based on validation performance. Fig.~\ref{fig:model_architecture} illustrates the overall classification pipeline.

\subsection{Sensor Feature Importance}

\begin{figure*}
    \centering
    \includegraphics[width=1\linewidth]{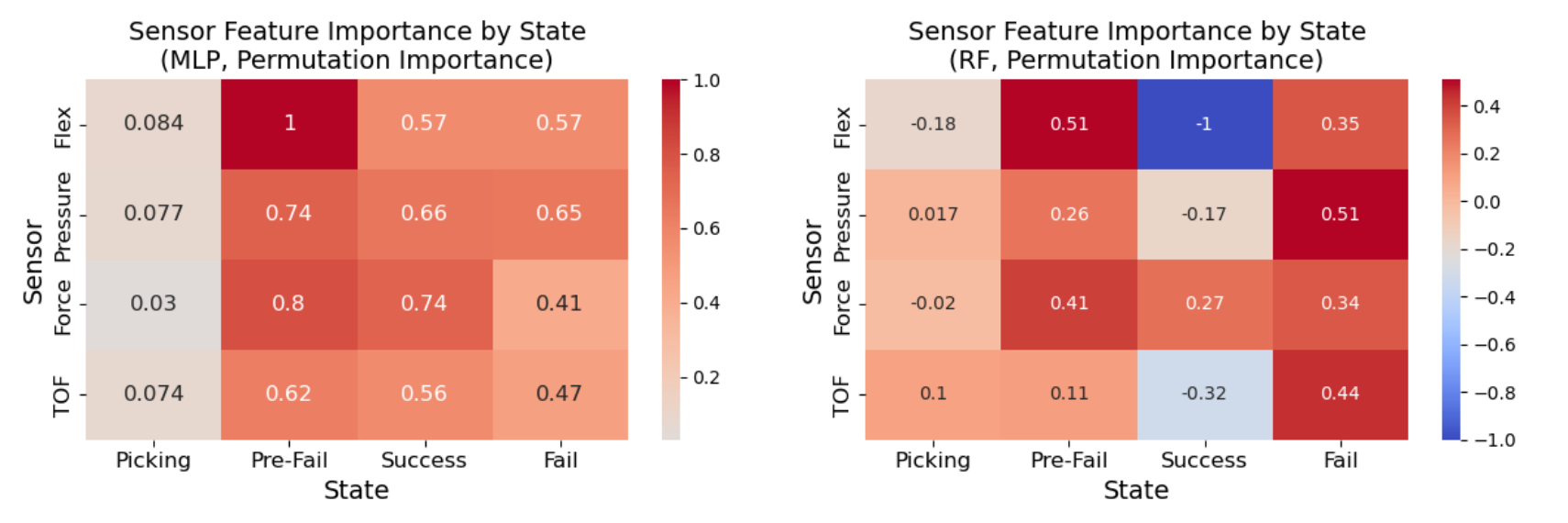}
    \caption{Permutation-based sensor importance across pick states for Random Forest and MLP models. Values are normalized within each state to highlight the relative contribution of force, vacuum pressure, flex, and time-of-flight (ToF) sensors during picking, pre-failure, successful pick, and failed pick phases.}

    \label{fig:heatmaps}
\end{figure*}

To quantify each sensor’s contribution, feature importance was computed from both models. For the Random Forest, importance scores for all time-step features were summed across time to yield one value per sensor. Permutation importance was also applied on the test set, measuring how shuffling each sensor affected classification performance for each pick state. The same permutation-based analysis was applied to the MLP. Results were visualized as state-by-sensor heatmaps to identify which modalities were most informative during different phases of the grasp (Fig.~\ref{fig:heatmaps}).

\subsection{Abscission Detection Evaluation Metrics}
\label{subsec:metrics}

Model performance was evaluated on the held-out test set using overall accuracy and class-specific precision, recall, and F1 scores for each pick state. These metrics quantify the system’s ability to detect both successful abscission and impending failure. Sensor relevance was assessed using the feature importance procedure described above.

\begin{figure}
    \centering
    \includegraphics[width=1\linewidth]{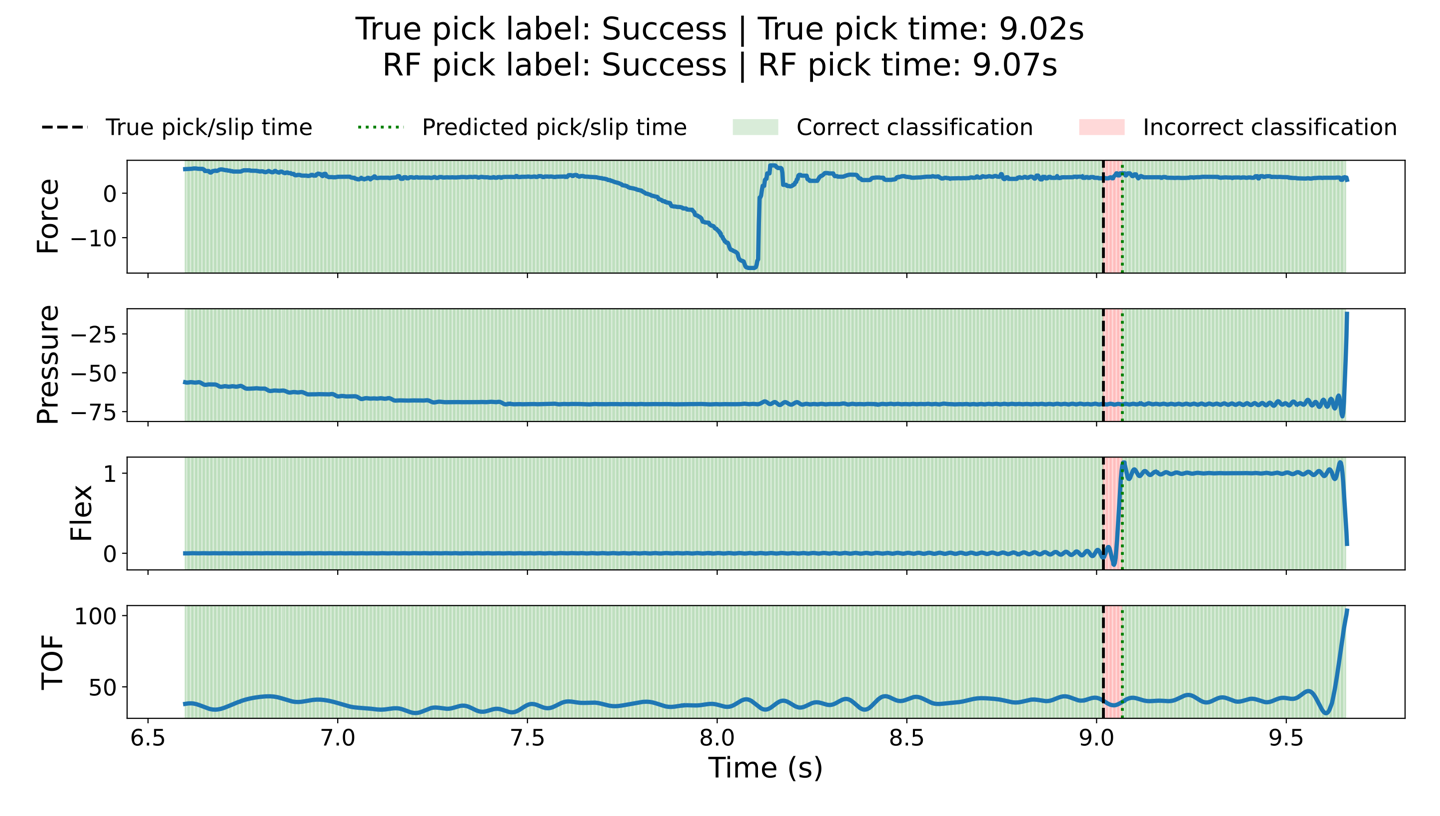}
    \caption{Example sensor time series from a single grasp trial. From top to bottom: filtered force, pressure, flex, and time-of-flight (TOF) signals aligned in time. Shaded regions indicate Random Forest (RF) classification windows, colored green when the window prediction matches the ground-truth label and red when it is incorrect. The dashed black vertical line marks the ground-truth pick or slip event, while the dotted vertical line marks the RF-predicted pick or slip time.}
    \label{fig:placeholder}
\end{figure}

\section{Results and Discussion}

\subsection{Classification Performance}
Both classifiers achieved high pick state classification accuracy under real orchard conditions. 
The Random Forest achieved a test accuracy of 91.3\%, while the MLP achieved 90.6\%, demonstrating that reliable multi-state grasp classification is feasible using only local sensing at the end-effector.

\begin{table}[!h]
\caption{Per-class precision, recall, and F1 score on the held-out test set for Random Forest (RF) and MLP pick state classifiers.}
\label{tab:per_class_performance}
\centering
\renewcommand{\arraystretch}{1.2}
\begin{tabular}{l|ccc|ccc}
\hline
 & \multicolumn{3}{c|}{\textbf{Random Forest}} & \multicolumn{3}{c}{\textbf{MLP}} \\
\textbf{State} & Prec. & Recall & F1 & Prec. & Recall & F1 \\
\hline
Picking      & 0.924 & 0.935 & 0.930 & 0.970 & 0.885 & 0.926 \\
Pre-failure  & 0.820 & 0.803 & 0.811 & 0.760 & 0.947 & 0.843 \\
Picked       & 0.922 & 0.973 & 0.947 & 0.921 & 0.934 & 0.927 \\
Failed pick  & 0.949 & 0.891 & 0.919 & 0.891 & 0.903 & 0.897 \\
\hline
\end{tabular}
\end{table}

Table~\ref{tab:per_class_performance} summarizes per-class performance. Both models performed strongest on the \emph{picking} and \emph{picked} states, corresponding to stable attachment and successful abscission, respectively. The \emph{pre-failure} state was more challenging, reflecting its transitional nature and limited temporal extent. Nonetheless, both models achieved high recall for pre-failure detection (MLP: 94.7\%, RF: 80.3\%), indicating that impending failures can be detected before complete loss of grasp.

Example sensor time series for a single grasp trial, along with Random Forest predictions aligned to ground-truth pick/slip events, are shown in Fig.~\ref{fig:placeholder}. Correctly classified windows are shaded green, while misclassified windows are shaded red, illustrating both the temporal alignment of predictions and the types of errors encountered.

Another metric used to evaluate classification performance was the time difference between the model’s predicted event time and the mean human annotation time. The random forest (RF) classifier predicted pick/slip events with an average error of 0.09 s relative to ground truth, which is well within the human annotators’ standard deviation (0.16 s). In contrast, the MLP exhibited a much larger average time error of 0.48 s. Although the MLP achieved high overall accuracy, it frequently produced early phase misclassifications during the picking process, leading to large timing errors despite correct final-state classification. The RF classifier was more robust to these early misclassifications.

\subsection{Sensor Importance Across Pick States}

Fig.~\ref{fig:heatmaps} summarizes sensor importance across pick states for both models using permutation-based analysis. While overall trends are consistent, important differences emerge between sensor modalities and grasp phases.

Across all states, flex and vacuum pressure sensors contributed strongly to classification performance in both models. In the MLP, flex sensors exhibited the highest overall importance (0.46), followed closely by pressure (0.44), indicating that local deformation and suction dynamics encode rich information about grasp quality. The Random Forest showed a similar global ranking, with flex contributing the largest aggregate importance (0.32), followed by pressure, force, and ToF.

State-wise analysis reveals that the \emph{pre-failure} state is strongly multi-modal. In both models, flex, force, and pressure sensors exhibited elevated importance prior to failure, suggesting that incipient slip manifests as a combination of subtle cup deformation, force redistribution, and suction degradation. This supports the use of sensor fusion for early failure detection rather than reliance on a single sensing modality.

In contrast, detection of the \emph{picked} state relied most heavily on force sensing, particularly in the MLP, where force exhibited the highest state-specific importance. This reflects the physical signature of successful abscission: once the stem detaches from the tree, the fruit load is fully transferred to the gripper and wrist, producing a distinct and sustained change in interaction forces.

While flex and pressure sensors also contributed to successful pick detection, their role was secondary. In the Random Forest analysis, flex features exhibited reduced or negative importance for the \emph{picked} state, likely due to increased signal variability after abscission as the compliant cup relaxes and deforms. In this regime, force sensing alone is often sufficient to determine pick success, and additional deformation signals may introduce noise or redundant information rather than improving discrimination.

\subsection{Random Forest vs. MLP Comparison}
While overall accuracy was comparable between models, qualitative differences emerged. The Random Forest demonstrated more balanced precision across states and fewer confusions between successful and failed picks, whereas the MLP achieved higher recall on the pre-failure class at the cost of slightly increased false positives. The Random Forest’s interpretability and consistent performance across sensor combinations make it well-suited for real-time deployment, particularly when paired with feature importance analysis for sensor selection. The MLP, however, appears better at capturing subtle temporal patterns preceding failure, suggesting potential benefits when early intervention is prioritized.

\subsection{Minimal Sensor Sets and Practical Implications}
Evaluating multiple sensor combinations revealed that performance improved substantially when local sensing modalities were added beyond wrist force sensing alone. Force sensing was the most reliable indicator of successful abscission, but performed poorly in isolation for detecting pre-failure states.

Combinations including flex and vacuum pressure sensors retained the majority of full-system performance for early failure detection, while the addition of force sensing improved confirmation of successful picks. Time-of-Flight sensing provided complementary benefits for failure verification but was less critical for predictive classification. Notably, while flex sensing is critical for detecting changes in contact stability prior to failure, its contribution diminishes after abscission, where deformation signals may become noisier due to compliant relaxation and dynamic motion of the detached fruit.

These results suggest a phase-dependent minimal viable sensing strategy: deformable flex and pressure sensing are most informative prior to failure, while force sensing is dominant for confirming successful picks. Importantly, all sensors used are low-cost, lightweight, and compatible with compliant suction-based grippers, making the proposed sensing suite suitable for agricultural robots operating under strict payload and cost constraints.

\section{Conclusion and Future Work}

This work demonstrates that reliable pick state classification for robotic apple harvesting can be achieved using local sensing integrated directly into a compliant suction-based gripper. In particular, Random Forest and MLP classifiers achieved high accuracy under real orchard conditions, with the Random Forest predicting pick or slip events within 0.09 seconds of human-annotated ground truth.

A key finding is that sensor relevance is phase-dependent. Deformable flex and vacuum pressure sensors are most informative during pre-failure phases, enabling timely detection of incipient slip. In contrast, wrist force sensing dominates successful pick detection by capturing the load transfer associated with stem abscission. This supports a minimal sensor strategy in which flex and pressure sensors provide early failure detection, augmented by wrist force sensing for robust abscission confirmation.

The contributions of this paper are twofold: we provide a phase-dependent evaluation of multimodal sensors for pick state classification in suction-based fruit harvesting, and we identify minimal sensor sets that maintain reliable detection under realistic orchard conditions. These findings enable low-cost, lightweight, and mechanically compatible sensing solutions suitable for deployment on agricultural robots. Future work will focus on integrating these insights into closed-loop control policies for real-time corrective actions and generalizing the approach across different crops, gripper designs, and orchard environments.

\bibliographystyle{IEEEtran}
\bibliography{citations}

\end{document}